\documentclass[a4paper,12pt,english,notitlepage]{article}
\usepackage{setspace}
\usepackage{fixltx2e}
\usepackage{graphicx}
\usepackage{babel}
\usepackage[cdot,squaren]{SIunits} 
\usepackage[small]{caption}

\usepackage{amsmath}    
\usepackage{amsfonts}
\usepackage{amssymb}
\usepackage{amsthm}

\usepackage{dcolumn}
\usepackage{bm}
\usepackage{wasysym} 
\usepackage{ulem} 
\usepackage[usenames,dvipsnames]{color}

\usepackage{subfig}
\usepackage{nicefrac}

\usepackage{pifont}
\usepackage{authblk}
\newcommand{\cmark}{\ding{51}}%
\newcommand{\xmark}{\ding{55}}%

\newcommand{\gait}[1]{\operatorname{\mathcal{#1}}}
\newcommand{\TO}{{\tiny\mars}}

\begin{document}

\title{Robustness: a new SLIP model based criterion for gait transitions in bipedal locomotion}

\author[1]{Harold Roberto Martinez Salazar}

\author[2]{Juan Pablo Carbajal}%
\author[3]{Yuri P. Ivanenko}

\affil[1]{%
 Artificial Intelligence Laboratory, Department of Informatics\\
University of Zurich, Switzerland. {\tt\small martinez@ifi.uzh.ch}
}%

\affil[2]{%
Department of Electronics and Information Systems\\
Ghent University, Belgium. {\tt\small juanpablo.carbajal@ugent.be}
}%

\affil[3]{%
 Laboratory of Neuromotor Physiology\\
 Fondazione Santa Lucia, Italy. {\tt\small y.ivanenko@hsantalucia.it}
}

\date{\today}

\maketitle
\begin{abstract}
Bipedal locomotion is a phenomenon that still eludes a fundamental and concise mathematical understanding. Conceptual models that capture some relevant aspects of the process exist but their full explanatory power is not yet exhausted. In the current study, we introduce the robustness criterion which defines the conditions for stable locomotion when steps are taken with imprecise angle of attack. Intuitively, the necessity of a higher precision indicates the difficulty to continue moving with a given gait. We show that the spring-loaded inverted pendulum model, under the robustness criterion, is consistent with previously reported findings on attentional demand during human locomotion. This criterion allows transitions between running and walking, many of which conserve forward speed. Simulations of transitions predict Froude numbers below the ones observed in humans, nevertheless the model satisfactorily reproduces several biomechanical indicators such as hip excursion, gait duty factor and vertical ground reaction force profiles. Furthermore, we identify reversible robust walk-run transitions, which allow the system to execute a robust version of the hopping gait. These findings foster the spring-loaded inverted pendulum model as the unifying framework for the understanding of bipedal locomotion.
\end{abstract}


\section{\label{sec:level1}Introduction}

The study of bipedal locomotion has motivated the development of several models that explain the most important principles governing the dynamics of the observed gaits. Some researchers have adopted models that include detailed representations of different leg components or that emulate neuromuscular structures using physical elements such as springs, dampers and multi-segmented legs. Although these models reproduce the dynamics of locomotion, their use as conceptual models is not widespread due to their complexity. In contrast, simpler models have been used extensively as conceptual models of bipedal locomotion~\cite{HolmesSIAM.06}.

Most of these simple models were developed to explain the exchange of kinetic and potential energy of the center of mass (CoM) of biological agents. During walking, kinetic and potential energy of the CoM are out of phase, i.e. the maximum height of the CoM corresponds with a minimum of its speed~\cite{Cavagna1977}. In consequence, the inverted pendulum (IP) model~\cite{Mochon1980} is frequently used to represent walking, since in this model the exchanges of energy are also out of phase. Detailed analyses of the passive dynamics of the IP model constituted a conceptual cornerstone for the development of mechanical devices capable of stable walking without any actuators or controllers~\cite{Collins2001}. Despite its conceptual explanatory power, the IP model does not correctly reproduce several aspects of human walking~\cite{Full1999}, e.g. the vertical oscillations of the CoM experimentally observed are smaller than the ones predicted by the model. Inspired in this model Srinivasan and Ruina proposed a biped model with ideal actuators on the legs~\cite{srinivasan2005}. They determined the periodic gaits that minimized the work cost assuming that the leg forces are unbounded if necessary. They found that transitions from walking to running at constant Froude number and step length are possible only when the Froude number is one. As a result, they found  an optimal walking gait that resembles the conditions of the walking gait at human walk to run transition, but at this condition they did not found an optimal running gait. In contrast, they identified a hybrid gait called pendular running which is not supported with the experimental data of human gait transitions. Further more, in this study the double support phase in walking was not allowed.  



Running is commonly represented with another model, the spring-loaded inverted pendulum (SLIP)~\cite{Blickhan1989}. The SLIP model consist of a point mass (the body) attached to a massless spring (the leg). During the stance phase the spring is fixed to the ground via an ideal revolute joint that is removed during flight phase. This model has been successfully used for the control of running machines~\cite{andrews2011}. In terms of combining multiple gaits, the explanatory power of the SLIP model surpasses that of the IP model, since the former can be extended to reproduce the mechanics of human walking by adding an extra massless spring representing the second leg, therefore unifying walking and runnig in a single model. However, the analyses carried out with the SLIP model had not yet explained gait transitions at constant forward speed, e.g. from walking to running at a characteristic Froude number. Previous studies suggested that transitions were only possible if the total energy was drastically increased or decreased to induce a considerable change in the forward speed of the system~\cite{Geyer2006}. With a simulation study~\cite{srinivasan2011}, Srinivasan explained gait transitions for springless bipeds model as a mechanism to minimize the energetic cost of the locomotion. However, in the case of springy biped systems the walk to run transition is not predicted by work minimization because for a certain range of stiffness it is possible to find work-free running at very low speeds.   


Given that the legs in the SLIP model are massless, their swinging motion cannot be directly described using equations derived from Newton's laws. Therefore, a control policy that sets the angle of attack at touchdown (the angle spanned by the landing leg and the horizontal at the time the foot collides with the ground) must be defined a priori. Generally, the angle of attack at touchdown is kept constant. Herein, we assume a more general control policy: the system selects a new angle of attack at each step. The study of the system is based on a return map. With the return map, we can understand the evolution of the dynamical system as a function of the selection of the gait and the angle of attack. This analysis is similar to~\cite{ernst2012,ernst2009,seyfarth2002b}, but in our study we define the return map at midstance. With this analysis, we can identify the initial conditions that, under this control policy, can perform a gait indefinitely. Instead of adding perturbations to the terrain to measure the robustness of the system as in~\cite{byl2009}, we extended the concept of viability introduced in~\cite{Martinez2011}, and assume that all the initial conditions with a valid control policy must be able to select an angle of attack inside a range of an arbitrary minimum size. We considered the length of the range of valid angles of attack as a qualitative measure of the robustness. The regions in which this control policy is valid are called robust regions, and regions where the system can change from one gait to another are called transition regions.


In this study, we propose this definition of robustness as a criterion to explain the onset of gait transitions, complementing the classical energetic criterion~\cite{Alexander1989,Minetti994}. Intuitively, the robustness of a gait can be understood as inversely related to the attentional demand required to maintain it. If highly precise inputs are needed to continue with a gait the system must spend more resources to select an adequate action, e.g. use of detailed models, better estimation of states from noise sensory data, more processing time; i.e. cognitive load or attention.

This new perspective is accompanied with a trade-off between robustness and energetic cost. A similar trade-off have been observed in bees~\cite{Combes2009}: when flying in turbulent flows, the animal extends its lower limbs reducing the chances of rolling, but increases the drag force sacrificing forward speed. Furthermore, the transitions found under the newly included robustness criterion qualitatively reproduce experimental values of the changes in the amplitude of the oscillations of the hip, changes in the gait duty factor and variations of ground reaction forces. Incidentally, these transitions use a gait pattern that we identify with hopping.

This paper is organized as follows. In section~\ref{sec:MatandMethRSB}, we define the models used for the simulation and introduce several concepts required for the understanding of the results. In section~\ref{sec:ResultsRSB} we show the regions of robust locomotion and gait transition. In that section we also compare our results with biological data. Discussions are given in section~\ref{sec:discussionRSB} and we conclude the paper in section~\ref{sec:conclusionRSB}.

\section{Definitions\label{sec:MatandMethRSB}}

The time evolution of a gait is segmented in several phases, each phase is described with a sub-model. These sub-models represent the motion of a point mass under the influence of: only gravity (flight phase), gravity and a linear spring (single stance phase),  gravity and two linear springs (double stance phase). The point mass stands for the body of the agent and the massless linear springs model the forces from the legs. During walking, running and hopping the system always goes through the single stance phase, therefore all gaits can be studied and compared during this phase. We denote the maps defined by walking, running and hopping as $\gait{W}$, $\gait{R}$ and $\gait{H}$, respectively. Given an initial state $x_i$ of the model, a walking step taken with angle of attack $\alpha$ is denoted $x_{i+1} = \gait{W}_\alpha(x_i)$ and similarly for running. As explained later a step of the hopping gait requires two angles, therefore it can be denoted with $x_{i+1} = \gait{H}_{\alpha\beta}(x_i)$.

The state of the system is observed when its continuous trajectory passes through a section, called $\mathcal{S}$. This section is defined by the support leg forming a right angle with the ground. At this section the state of the system is defined by the height of the hip (i.e. height of the CoM), $r$, and the velocity in the vertical direction, $v_y$ (see Appendix~\ref{ap:equations} for more details).

All initial conditions are given in the $\mathcal{S}$ section and in the single stance phase, i.e. only one leg touching the ground and oriented vertically. $(r,v_y)$ pairs were simulated for values of the total energy $E$ in the range $[780,900]\joule$ at intervals of $\unit{10}\joule$. The model was implemented is in MATLAB(2009, The MathWorks) and simulations were run using the step variable integrator ode45. Experimental data analysis was performed using GNU Octave.

\subsection{Viability, Robustness, symmetric gaits and biomechanical observables\label{sec:viabilityRobustRSB}}
{\it Viability}, as presented in~\cite{Martinez2011}, defines the easiness of taking a further step during locomotion. That is, the wider the range of angles of attack that can be used to take a step the easier is to take that step. In a physical platform it is required that a valid angle of attack exists for a definite interval, since real sensors and actuators have a finite resolution and are affected by noise. A viability region in the section $\mathcal{S}$ contains all the states for which at least one step can be taken selecting an angle of attack from an interval of at least $\Delta \alpha$, i.e. states for which if at least one iteration of the gait is applied map into states of the same gait. For example, for the running gait, this can be expressed as,
\begin{equation}
\begin{split}
V^R\left(\Delta\alpha\right) = & \lbrace x \vert \; x \in \mathcal{S} \wedge \\ & \left( \exists \alpha \in I_\alpha, \; \Vert I_\alpha \Vert \geq \Delta\alpha \; \vert\; y = \gait{R}_\alpha \left(x\right), \; y \in \mathcal{S} \right)\rbrace.
\end{split}
\end{equation}

\noindent Where $I_\alpha$ stands for the angle interval and $\Vert I_\alpha \Vert$ for its size. Narrower angle intervals, i.e. more precise angle definition, lead to bigger viability regions and wider intervals to smaller regions. An example of the viability regions can be found in appendix~\ref{ap:equations}.

The concept of {\it robustness} is defined on top of that of viability. A state in the robust region is a viable state that can always be mapped into the robust region by choosing the appropriate angle of attack. This angle should be viable, i.e. it must be selected from an interval of at least $\Delta \alpha$. For example, for the walking gait, this can be expressed as,
\begin{equation}
\begin{split}
\rho^W\left(\Delta\alpha\right) = & \lbrace x \vert \; x \in \rho^W\left(\Delta\alpha\right) \wedge \\ & \left( \exists \alpha \in I_\alpha, \; \Vert I_\alpha \Vert \geq \Delta\alpha \; \vert\; y = \gait{W}_\alpha \left(x\right), \; y \in \rho^W\left(\Delta\alpha\right) \right)\rbrace.
\end{split}
\end{equation}

\noindent Where $I_\alpha$ stands for the angle interval and $\Vert I_\alpha \Vert$ for its size. This assumes that the controller can select an angle of attack for each step. In particular, this includes constant angle of attack policies and some of the self-stable regions identified in~\cite{Geyer2006} belong to a robust region. However, this does not mean that the system remains in the self-stable region for each step, since that would imply that the angle of attack is selected precisely. Instead, robustness implies that if the system was in that region at time $t$, it can remain close to it, even if the angles are selected with finite resolution.

The gaits commonly used by humans are symmetric, meaning that the dynamical behavior of the left leg mirrors the one of the right leg. In our model this is possible when two conditions are satisfied: the velocity in the vertical direction at $\mathcal{S}$ is zero and there is an angle of attack $\alpha$ that can bring the system back to the same state.

In the subsequent section we will show that the discovery of robustness as a useful criterion to induce gait transitions allows for qualitative comparisons with experimental biomechanical data. In particular we present results in terms of {\it Froude number}, {\it hip excursion}, {\it gait duty factor}, and {\it vertical ground reaction forces}. The Froude number is the ratio between the weight and the centripetal force $\nicefrac{w^2l_o}{g}$, where $g$ is the acceleration due to gravity, $l_o$ is the natural length of the leg and $w$ is the angular velocity of the body around the foot in contact with the ground. Hip excursion denotes the amplitude of vertical oscillations of the hip. The gait duty factor is the fraction of the total duration of a gait cycle in which a given foot is on the ground. The vertical ground reaction force is vertical component of the normal force exerted by the ground.

\section{Results\label{sec:ResultsRSB}}
We report the results obtained from the study of gait transitions in the SLIP model following the criterion of robustness detailed in Section~\ref{sec:viabilityRobustRSB}. It turns out that the concept of robust gaits offer an alternative explanation for the onset of gait transitions in bipedal locomotion, comparable with arguments based on metabolic costs.

We begin our exposition with a detailed explanation of the conditions, in terms of decrease of robustness, that may trigger gait transitions. From there we move on to describe the mechanism underlying robust gait transitions. The results of those two sections are combined to present qualitative comparison with biomechanical observables, followed by a short description of robust hopping.

\begin{figure*}[htb]
\centering
\includegraphics[width=0.98\textwidth]{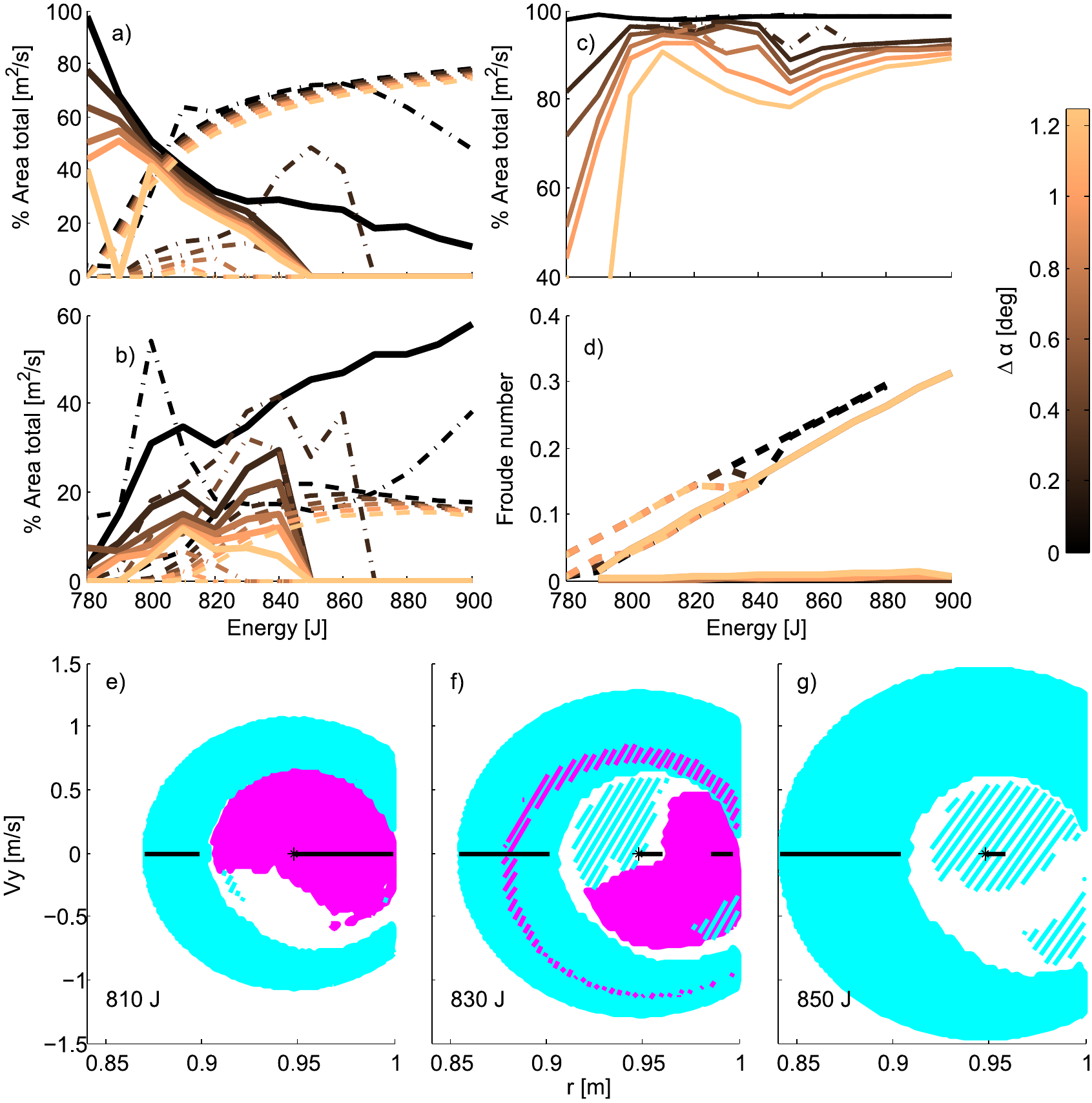}
\caption{\label{fig:VxVblStbl} (Color online) Robust regions. For panels {\bf (a)} - {\bf (d)} the (copper) gray color scale represents the interval size used to calculate the robust region. {\bf (a)} shows the robust region area in the section $\mathcal{S}$ for running (dashed line), walking (continuous line), and grounded running (dash-dotted line). {\bf (b)} shows the area of viable transitions that brings the system to robust running (dashed line), robust walking (continuous line), and robust grounded running (dash-dotted line) in the section $\mathcal{S}$. {\bf (c)} shows the total area in the section $\mathcal{S}$ cover by the robust gaits and the viable transitions. The dash-dotted line represents all the gaits, and the continuous line represents walking and running. {\bf (d)} shows the maximum and minimum Froude number for a robust gait at the section $\mathcal{S}$ for different energies.  Robust walking is depicted with the dashed line, and robust running is depicted with the continuous line. In panels {\bf (e)} - {\bf (g)} filled patches represents robust running ((blue) light gray) and robust walking ((magenta) dark gray) in the section $\mathcal{S}$. The dashed region represents viable transition to robust running using walking ((blue) light gray), and to robust walking using running ((magenta) dark gray). The solid black line depicts the symmetric gaits.}
\end{figure*}

The definition of robust gait applies for symmetric and non-symmetric gaits. Figure~\ref{fig:VxVblStbl}a shows the area of the robust regions in the section $\mathcal{S}$ for different energies and different interval lengths $\Delta \alpha$. With this model we identify three different gaits: running, walking and grounded running. Grounded running has the same phases as walking but in the transition from the single support to the double support the vertical velocity of the center of mass is positive while in walking the velocity is negative (Appendix~\ref{ap:equations}). Results show that the grounded running gait is less robust that walking and running. For a $\Delta \alpha$ bigger than $\unit{0.5}\degree$, the grounded running gait covers less than 15\% of the initial conditions in the section $\mathcal{S}$.

Figure~\ref{fig:VxVblStbl}b shows the area of the viable transitions to the robust regions in the section $\mathcal{S}$ for different energies and different interval lengths $\Delta \alpha$. For example, the viable transition to robust running considers the initial conditions outside robust running that under walking or grounded running can be brought to robust running in one step. Given that this transitions are viable the angle of attack can be selected from an interval of length $\Delta \alpha$. A similar condition is applied to calculate the viable transition to robust walking or robust grounded running. For a $\Delta \alpha$ bigger than $\unit{0.75}\degree$, the viable transition to robust grounded running gait covers less than 10\% of the initial conditions in the section $\mathcal{S}$. Figure~\ref{fig:VxVblStbl}c shows the total area of robust regions and viable transitions with and without grounded running. Results show that for a $\Delta \alpha$ bigger than $\unit{0.5}\degree$ grounded running does not cover different initial conditions from walking and running.

Figure~\ref{fig:VxVblStbl}d shows the range of forward speed for robust running and walking at several energies and different interval lengths $\Delta \alpha$. Results show that the length of the interval affects the maximum Froude number in the walking gait. The bigger the $\Delta \alpha$, the lower the walking Froude number. In addition considering an interval length lower than $\unit{1} \degree$, robust walking exists only at low locomotion energies, while running increases robustness for higher energies. For an interval length bigger than $\unit{1} \degree$ walking  walking is not possible in all the low energy levels.  

We can draw an analogy between the results of the system with an interval length lower than $\unit{1} \degree$ and the experimental results reported in~\cite{Abernethy2002}, where it was shown that imposed fast walking required higher attention than running at similar speeds. Furthermore, normal switching between gaits did not required high attentional demand.


\subsection{Conditions for transitions\label{sec:transcond}}

We studied the transitions for a robustness criterion of $\Delta \alpha$ equal to $\unit{1} \degree$ because this was the limit condition in which the results of attentional demand can be qualitatively explained by the model. In addition we focused in the walking and running gait given that grounded running does not provide new possible states from the ones identified in robust walking and robust running (Fig.~\ref{fig:VxVblStbl}c). All the possible states of the system in the section $\mathcal{S}$ lie in a hemispherical region (see equations (15)-(21) of~\cite{Martinez2011} and Appendix~\ref{ap:equations}). In Fig.~\ref{fig:VxVblStbl}e-g, we marked the apex of this hemisphere with a star symbol. The closer the system is to the star, the higher the forward speed of the gait. Symmetric gaits are marked with a solid line, all symmetric gaits have $v_y=0$. The figure shows that symmetric robust walking moves away from the apex of the hemisphere as energy increases, i.e. it becomes slower. At $\unit{830}\joule$ symmetric robust walking is constrained to the rightmost side of the viability region reducing the speed of this gait considerably. Furthermore, at this energy the region of symmetric walking breaks down into two unconnected segments. This is also evident in Fig.~\ref{fig:VxVblStbl}d where the maximum speed of symmetric robust walking shows a strong slowdown with a sudden change of slope. The latter is a consequence of the rupture of the symmetric gait region. This milestone in the evolution of the gait can be used as a natural trigger for a gait transition.


The evolution of the area of robust walking, and robust running, are shown in detail in Figure~\ref{fig:VxVblStbl}e-f. This figures show that, at low energy, robust walking covers a wide region of the viable states of the system, while at high energy robust running covers a wider area. Around $\unit{800}\joule$ both robust gaits have similar area. Based on robustness alone, this will imply a transition. However, symmetric robust walking intersects the apex of the hemisphere producing the fastest forward speed up to energies of $\unit{810}\joule$, favoring walking in terms of energy efficiency. When the energy is increased further, the area of robust walking decreases and symmetric robust walking is constrained to low speeds. Due to these facts, at energies close to $\unit{840}\joule$, the speed of symmetric robust walking and running match. For higher energies the gait transition is imminent, since the only robust gait remaining is symmetric running.


\subsection{Mechanism of gait transitions\label{sec:transmech}}
Assuming that during locomotion the fastest robust gait patterns are preferred over slower or non-robust ones, we see that for energies below $\unit{840}\joule$ walking is the gait of choice and for energies above that value running would be chosen. Therefore, we study viable transitions at $\unit{840}\joule$ and compare them with results from an experiment on human gait transition. We consider transitions only when all angles of attack used in the process can be chosen from an interval of length $\unit{1}\degree$ or greater, i.e. we define admissible transitions using the concept of viability (sec.~\ref{sec:viabilityRobustRSB}).

We consider two mechanisms to execute gait transitions between symmetric robust gaits (symmetric gaits are known to be self-stable and therefore a good choice for stable locomotion, see~\cite{Geyer2006}). The first mechanism, which can only be used from walking to running, consist in moving from the robust region of walking to the viability (non-robust) region of the same gait, and from there select an angle of attack to go to the robust region of running. This mechanism  can be used in robust walking between $\unit{830}\joule$ and $\unit{840}\joule$ (see Figure~\ref{fig:TrnstionRegion}a). The second mechanism  consist in going from a robust region of a given gait (walking or running) directly to the robust region of a different gait. This mechanism  is applicable for robust running between $\unit{830}\joule$ and $\unit{840}\joule$ while in robust walking is only applicable around $\unit{840}\joule$.

These mechanisms can be further constrained by selecting desired properties of the final gait. One possibility is to execute a transition in such a way that the final gait has the same (or as close as possible) Froude number as the initial gait. Another possibility is to execute a transition that sets the hip excursion of the new gait to a desired value (see Figure~\ref{fig:TrnstionRegion}b for a graphical description). These constraints are referred in this study as strategies and they are used for the comparison between our simulated results and experimental data presented in the next section.

\subsection{Qualitative Prediction of Biomechanical Observables\label{sec:Biomech}}

As we mention before, the biomechanical observables used to compare our results with experimental data are: Froude number, hip excursion, gait duty factor and vertical ground reaction forces. In the Appendix~\ref{ap:angleAttkEst}-\ref{ap:phaseHip}, we extended this comparison to include angle of attack sequences and change of phase. We compare all our simulations against the experimental data reported in Figure 2 of~\cite{Ivanenko2011}, we will refer to this data as ``experimental data'' or ``the experiment''.

\begin{figure*}[tbp]
\centering
\includegraphics[width=0.99\textwidth]{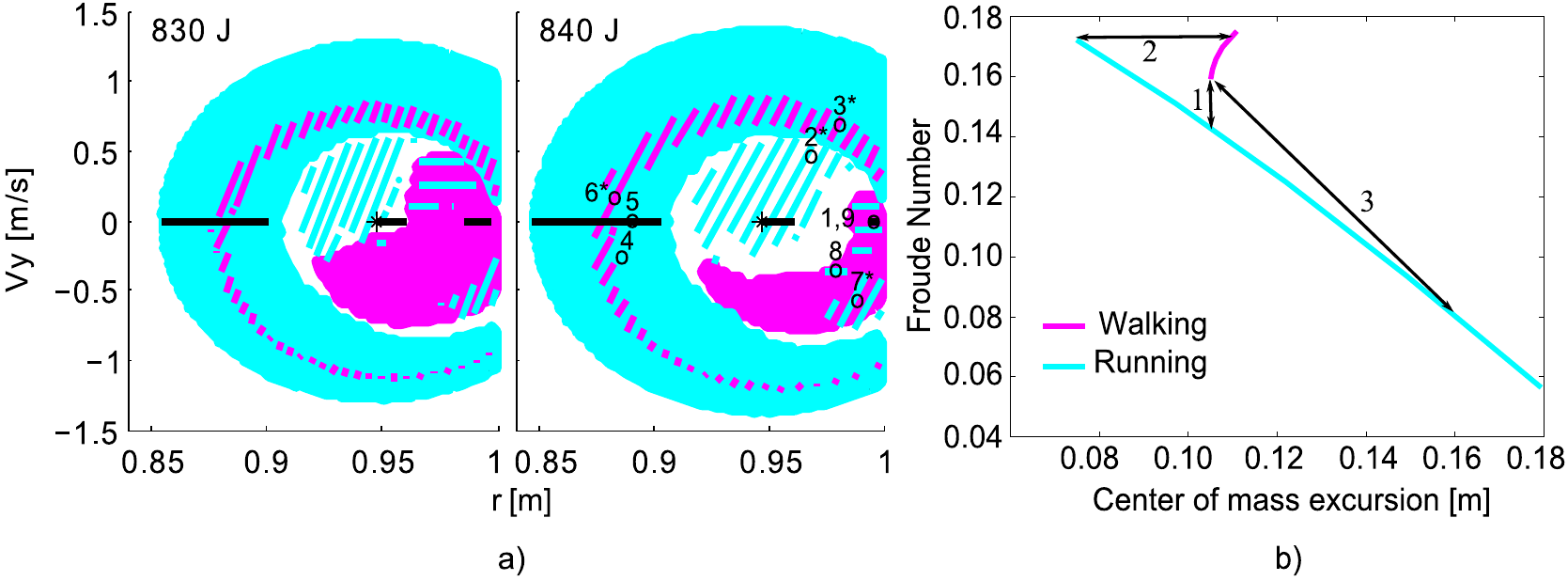}
\caption{\label{fig:TrnstionRegion} (Color online) Viable transitions. In all panels (blue) light gray color represents running and (magenta) dark gray color represents walking. {\bf (a)} shows viable transitions at two energy levels. Filled patches corresponds to robust regions. Shaded regions inside these are viable transitions regions. Diagonal shading corresponds to regions where the system can change between robust gaits (non-symmetric) in only one step. The horizontal shading delimits the region where the system can go to the non-robust transition region. The right panel shows two transition using both mechanisms. See text for details. {\bf (b)} shows the Froude number versus hip excursion for symmetric robust running and walking at $\unit{840}\joule$. Arrows indicate: (1) constant hip excursion, (2) constant Froude number and (3) relative change of the amplitude of the hip excursion fitted to experimental data.}
\end{figure*}

Figure~\ref{fig:TrnstionRegion}a shows the transition regions at two energy levels. We painted the robust regions of running and walking with a solid color, the shaded regions inside these are transitions regions where the system can change the gait. The diagonal shading corresponds to regions where the system can change between robust gaits (non-symmetric) in only one step. The horizontal shading delimits the region where the system can go to the non-robust transition region, as described in \ref{sec:transmech}. The right panel shows examples of a transition from walking to running and another from running to walking using the two mechanisms mentioned in the previous section. For the first transition, the system starts at symmetric robust walking (1), in the first step it moves to the non-robust transition region (2*) and executes the transition to robust running (3*). With two further steps the system is able to reach symmetric robust running (4-5). The transition in the other direction starts at symmetric robust running (5). Then the system moves to the robust transition region (6*) from which, in a single step, it changes to robust walking (7*). With two more steps the system reaches symmetric robust walking (8-9). In both transitions, the hip excursion was kept as constant as possible.

Figure~\ref{fig:TrnstionRegion}b shows the Froude number and the hip excursion of all symmetric robust gaits at $\unit{840}\joule$. As indicated in the figure, vertical transitions keep the hip excursion constant, while horizontal transitions produce gaits with the same Froude number.

\begin{figure*}[htbp]
\centering
\centering
\includegraphics[width=0.9\textwidth]{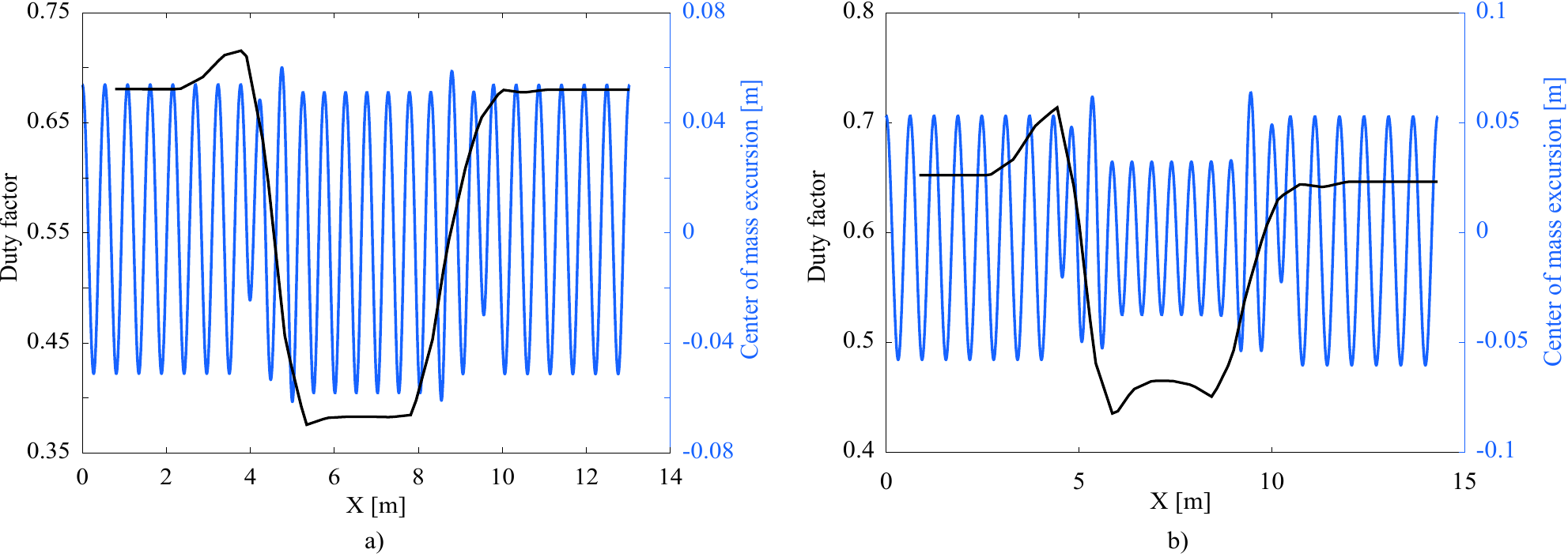}
\caption{\label{fig:DFE} (Color online) Hip excursion and gait duty factor for transition at constant hip excursion {\bf (a)}; and constant Froude number {\bf (b)}. The (blue) light gray color represents the hip excursion and the black line represents the duty factor. The plots show several steps before and after each transition.}
\end{figure*}

Figure~\ref{fig:DFE} shows time series of hip excursion and duty factor for a transition at constant hip excursion, together with a transition at constant Froude number. In both situations we obtain a Froude number that is about 60\% smaller than the one found in human gait transitions, which is around of 0.5~\cite{Ivanenko2011}. Nevertheless the SLIP model provides the best Froude number estimation to the date, when compared to other simple models, e.g. the IP model.


Ground reaction forces prior to the transition from walking to running have three main characteristics~\cite{Li2002}. Firstly, they present an asymmetric double bell-shaped profile. Secondly, the earlier peak becomes bigger than the later one and, thirdly the depression between the peaks becomes more accentuated in the last step of walking, exactly before the transition. In the case of the transition from running to walking, it was reported that the vertical ground reaction forces decrease during the steps prior to the transition.

\begin{figure*}[htb]
\centering
\includegraphics[width=0.9\textwidth]{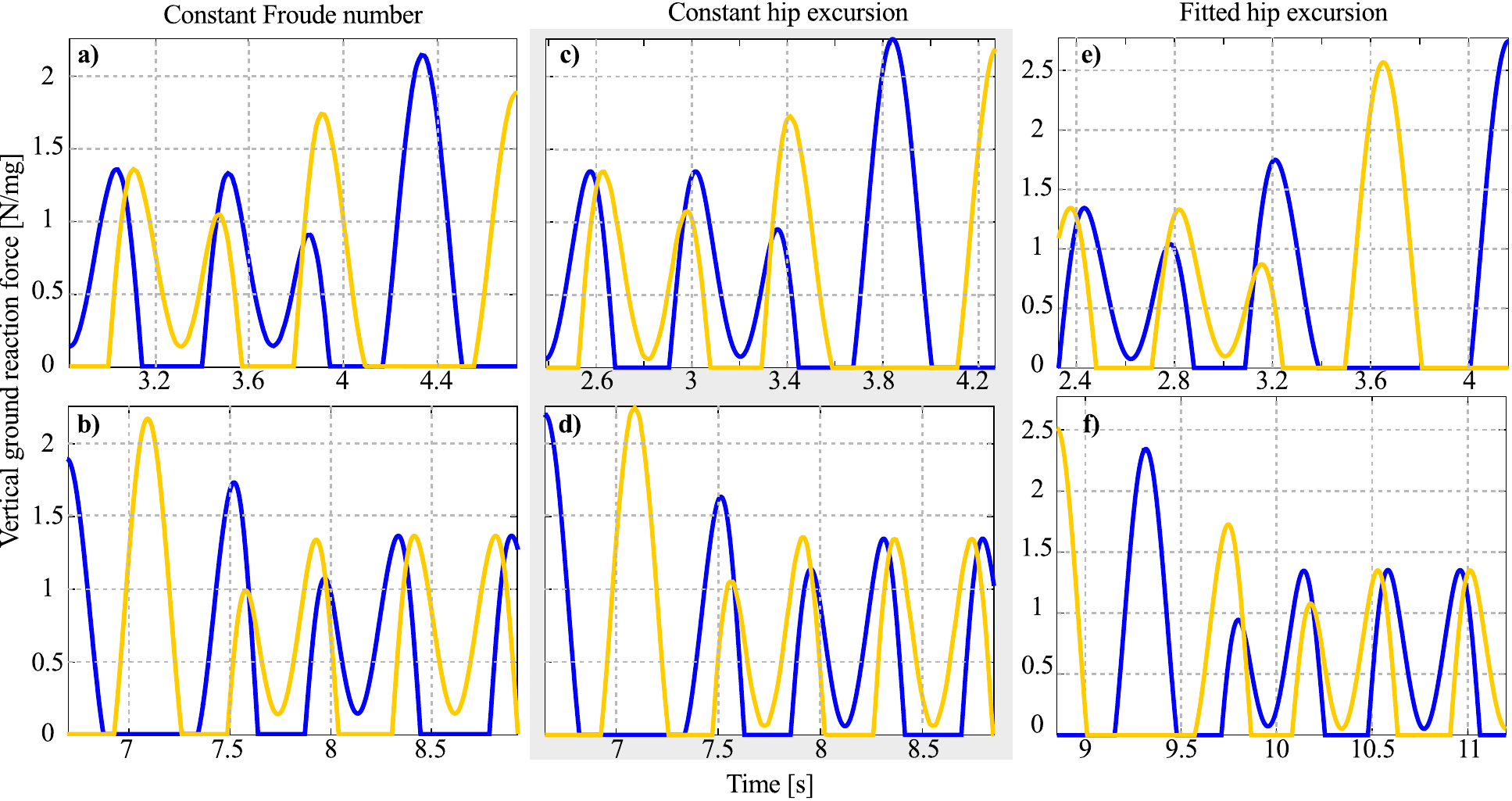}
\caption{\label{fig:GRFTransition} Vertical ground reaction forces during transitions. The six panels show a transition from symmetric robust walking to symmetric robust running with three different strategies, (a)-(b) constant Froude number, (c)-(d) constant hip excursion, (e)-(f) hip excursion similar to the experimental data. The forces present an asymmetric double bell-shaped profile. In the walking to running transition, (a)-(c) and (e), the earlier peak becomes bigger than the later one, exactly before the transition. The transitions in the other direction, running to walking (b)-(d) and (f) show vertical ground reaction forces that decrease considerably in the last running step due to the support of the second foot. The selection of a hip excursion similar to the experimental data introduces a progressive reduction of the force peak in more than one step (f). All forces are normalized with respect to the weight of the system.}
\end{figure*}

In Figure~\ref{fig:GRFTransition} we have plotted the vertical ground reaction forces for three different simulated examples. The first row of panels shows transitions from walking to running, and the second row of panels shows transitions in the other direction. Panels (a) and (b) show transitions keeping the Froude number constant. Panels (c) and (d) show transitions at constant hip excursion. The last example, presented in the panels (e) and (f), shows transitions that match the change in amplitude that was observed in the experiment. All cases qualitatively match the characteristics of the ground reactions reported in~\cite{Li2002}. The decrement in the force of the last running step is due to the support of the second foot. A reduction of the peak in more than one step appears only on the case where we matched the hip excursion of the experimental data.

In Table~\ref{tab:StrategiesComp}, we present a summary of the comparison between the simulated examples and the experimental data. Each column is discussed next.
Due to the variety of transitions that can be generated with the model, the number of steps to execute them can be select in a wide range, at least from 3 to 8 steps. From Figure~\ref{fig:TrnstionRegion}b we can see that the Froude number of all these transitions are lower than 0.5, this reflects the fact that the simulations have lower forward speeds ($v_x$) than the observed in humans. As pointed before, the many transitions that can be simulated, permit the matching of the relative change in hip excursion ($\Delta r$) measured in the experiment. In all simulated transitions the vertical ground reaction forces ($F_y$) are qualitatively well reproduced. The selection of the angle of attack are qualitative similar to what we found in the experimental case: the system moves progressively from one gait to the other changing the angle of attack at each step. However, the oscillation of the hip before and after the simulated transitions presents a change of phase ($\Delta \phi$) that not always coincide with what is observed in reality. Details for these two observables are presented in the the Appendix~\ref{ap:angleAttkEst}-\ref{ap:phaseHip}.


\begin{table*}[t]
\centering
\begin{tabular}{|l|c|c|c|c|c|c|}
\hline
Strategy             & \# Steps   & $v_x$ & $\Delta r$ & $F_y$ & $\Delta \alpha$ & $\Delta \phi$\\
\hline
Const. Froude number & \cmark     &\xmark & \xmark     &\cmark & \cmark   & \xmark       \\
\hline
Const. hip excursion & \cmark     &\xmark & \xmark     &\cmark & \cmark   & \xmark       \\
\hline
Fitting experiment   & \cmark     &\xmark & \cmark     &\cmark & \cmark   & \xmark       \\
\hline

\end{tabular}
\caption{\label{tab:StrategiesComp} Comparison between three transition strategies and experimental data. The symbol \cmark\, indicates qualitative matching between simulation and experiment, while the symbol \xmark\, indicates the opposite. $v_x$: forward speed of the center of mass; $\Delta r$: relative change in hip excursion before and after transition; $F_y$: vertical ground reaction forces; $\Delta \alpha$: change of the angle of attack during transition; $\Delta \phi$: change in phase of the oscillations of the hip before and after transition.}
\end{table*}

\subsection{Robust Hopping Gait\label{sec:HoppingGaitRSB}}

\begin{figure*}[htb]
\centering
\includegraphics[width=0.8\textwidth]{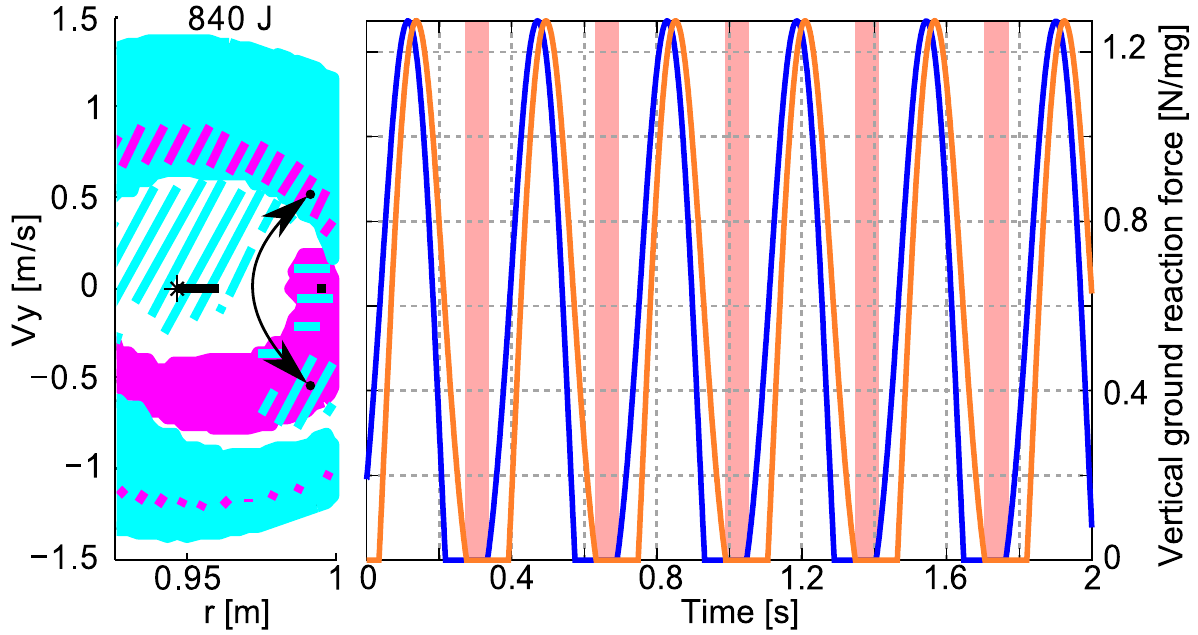}
\caption{\label{fig:GRFHopping} (Color online) Vertical ground reaction forces during hopping. Panel (a) shows the transition regions in section $\mathcal{S}$ for $E=\unit{840}\joule$; the arrows show the states in the robust transition region that are used alternately. Panel (b) shows the ground reaction forces for each leg. The (pink) gray rectangles show the different flight phases. The forces from the legs are indicated with solid lines with different colors.}
\end{figure*}

At $\unit{840}\joule$ we identify a transition region in robust walking where the system can go in one step to robust running. Among the states in this transition region, there a some that are mapped directly into the transition region of robust running. By selecting alternatively the right angles of attack, the system can sequentially walk and run, producing the hopping gait. Fig. \ref{fig:GRFHopping} shows an example of this gait. By looking at the vertical ground reaction forces in the figure, we see the different phases that compose this gait; from single stance phase to double stance phase then to single stance phase and finally to flight phase.

\section{Discussion\label{sec:discussionRSB}}

Herein we have modeled bipedal locomotion using the SLIP model. This model conserves the total mechanical energy and at first glance it may seem inapposite for the prediction of gait transitions, since work has to be done on the system to increase the speed of locomotion. Nevertheless, by looking at the behavior of the model at different energies, we can emulate the situation where work is done on the system.

We proposed robustness as a new measure of the easiness of locomotion. Robustness measures the level of attention that needs to be dedicated to take a step; the more robust a gait is, the less attention that is needed to take the next step.

According to our results, the selection of the gait can be based on two criteria: efficiency, which is the selection of the gait with the highest forward speed; and robustness, which defines how easy is to maintain the given gait. This second criterion is consistent with the experimental results of attentional demand in locomotion reported in~\cite{Abernethy2002}. Based on these criteria, walking is the best choice for energies below $840\joule$, and running is more appropriate for higher energies. This resembles what is observed in human locomotion.

Using robustness as the leading criterion, we identify transition regions that allow the system to go from one gait to the other even in the case of imprecise angle selection. These transition regions are present for energies from $\unit{830}\joule$ to $\unit{840}\joule$ (Fig.~\ref{fig:TrnstionRegion}a). At $\unit{840}\joule$, symmetric robust running and walking share all the possible velocities, facilitating gait transitions. In the case of an increment of energy, to keep robustness and move forward faster, a walking system can execute a transition to robust running at $\unit{840}\joule$. The transition can be reversed when the system decreases its energy. Note that the mechanisms of transition shown in Fig.~\ref{fig:TrnstionRegion}a (right panel), have the following properties. One mechanism connects the robust region of both gaits, while the other one connects the non-robust viability region of walking with robust running. The latter mechanism is not reversible, meaning that the system cannot go from running back to this region in a single step. The transitions connecting robust regions are reversible and the system can oscillate between the two gaits robustly. Is in this situation where the hopping gait emerges. This locomotion pattern is frequently used by children when playing joyfully.

The existence of non-empty transition regions (Fig.~\ref{fig:TrnstionRegion}b) implies that the system has multiple alternatives to change gaits. These alternatives will produce different changes of forward speed and hip excursion. We show three different scenarios: constant hip excursion, hip excursion similar to experimental data and constant Froude number.

When the transition matches the hip excursion of the experimental data, the Froude number varies from $0.16$ in walking to $0.08$ in running, while in the experiment it is almost constant (slowly varying treadmill speed, see~\cite{Ivanenko2011} for details on the experiment). As explained before, in all simulated cases the absolute values of Froude number are lower than in the experiments. The hip excursion has an amplitude of $\unit{5.2}\centi\metre$ in walking and $\unit{8.3}\centi\metre$ which also similar to the one reported in~\cite{Ivanenko2011} which is around $\unit{7}\centi\metre$.

When the transition keeps the Froude number constant the hip excursion decreases from $\unit{5.7}\centi\metre$ in walking to $\unit{3.7}\centi\metre$ in running. This contradicts the behavior observed in our experimental data. The simulated Froude number for this transition is about $0.17$.

The robustness criterion induces an underestimation of the forward speed at gait transitions. The highest Froude number achieved using the previous strategies is around one third of the one observed in humans ($0.5$). However, given the strong simplifications in the model the result is encouraging. To reduce the gap between simulated and experimental Froude number, the model can be extended to include the displacement of the point where the leg is in contact with the ground during the stance phase~\cite{Adamczyk2006}.

All transitions presented here produce similar results concerning the duty factor. Walking has a duty factor around $0.7$ and running has a duty factor around $0.4$, in accordance with the experiment. Furthermore, in all transitions from walking to running the model predicts a progressive change in the vertical component of the reaction forces, i.e. the relation between the first and the second peak of the force during the transition. This also applies to the transitions from running to walking. In particular, the ground reaction forces corresponding to transitions matching the hip excursion of the experimental data (Fig. \ref{fig:GRFTransition}) introduces a progressive reduction of the force peak in more than one step. All these results qualitatively reproduce the experimental results reported in~\cite{Li2002}.

\section{Conclusion\label{sec:conclusionRSB}}

The comparison between experimental data and simulations using the SLIP model shows that the model is not able to generate accurate quantitative predictions. Most strikingly, the forward speed in the simulations are considerable slower than that observed experimentally. This difficulty can be overcome by adding a more detailed description of the contact between leg and ground. Nevertheless, the SLIP model can be used as a conceptual model to explain the many aspects of bipedal locomotion such as the mechanics of running, walking, hopping and gait transitions.

Our findings indicate that robustness can play an important role in inducing gait transition, complementing the usual view focused solely in energy expenditure. The robustness criterion is analogous to the attentional demand during locomotion and may play an important role deciding the gait transition events. To our knowledge this is the first time such a criterion is included in a numerical model of locomotion.

\paragraph{Acknowledgements} The research leading to these results has received funding from the European Community's Seventh Framework Programme FP7/2007-2013-Challenge 2-Cognitive Systems, Interaction, Robotics- under grant agreement No 248311-AMARSi.

\paragraph{Authors contribution} {\bf HMS} developed the computational and mathematical model, run the simulations and performed data analysis. {\bf JPC} collaborated in development of the mathematical models, the data analysis and the interpretation of results. {\bf YI} collected and contributed the experimental data. All authors contributed to the writing of this manuscript.

\bibliographystyle{unsrt}
\bibliography{referencesRSB}

\appendix

\section{Equations of motion\label{ap:equations}}

We define a running gait as a trajectory that switches from the single stance phase to the flight phase and back to the single stance phase. A walking gait is defined as a trajectory that switches from the singles stance phase to the double stance phase and back again to the single stance phase.

The state in the flight phase is represented in Cartesian coordinates of the position of the point mass and its velocity $\vec{X}_{ff} = \left(x,y,v_x,v_y\right)^T$,
\begin{equation}
\dot{\vec{X}}_{ff}  = \begin{pmatrix}v_x\\ v_y\\0\\-g\end{pmatrix},
\label{eq:ff-chart}
\end{equation}

\noindent where $g$ is the acceleration due to gravity.

The state in the single stance phase is represented in polar coordinates $\vec{X_s} = \left(r,\theta,\dot{r},\dot{\theta}\right)^T$, where $r$ is the length of the spring and $\theta$ is the angle spanned by the leg and the horizontal, growing in clockwise direction. Thus, the equations of motion are:
\begin{equation}
\dot{\vec{X}}_{s}  = \begin{pmatrix}\dot{r}\\ \dot{\theta}\\ \frac{k}{m}\left( r_0-r\right)+ r\dot{\theta}^2-g\sin\theta\\ -\frac{1}{r}\left(2\dot{r}\dot{\theta} +g\cos\theta\right) \end{pmatrix}.
\label{eq:s-chart}
\end{equation}
\noindent It is important to note that $\theta(t_{TD}) = \alpha$, i.e. the angular state at the time of touchdown is equal to the angle of attack. The parameter $r_0$ defines the natural length of the spring.

In the double stance phase the state is also represented in polar coordinates $\vec{X_d} = \left(r,\theta,\dot{r},\dot{\theta}\right)^T$, with the origin of coordinates in the new touchdown point. The motion is described by:
\begin{eqnarray}
& \dot{\vec{X}}_{d}=\begin{pmatrix}
\dot{r}\\
\dot{\theta}\\
\begin{split}
\frac{k}{m}&[(r_0-r) + \left(1-\frac{r_0}{r_{\TO}}\right)\ldots \\ &(x_{\TO}\cos\theta - r) ]  +r\dot{\theta}^2 \ldots  \\ &-g\sin\theta
\end{split}
\\
\begin{split}
-\frac{1}{r}[&\frac{k}{m}\left(1-\frac{r_0}{r_{\TO}}\right)x_{\TO}\sin\theta \ldots \\ & + 2\dot{r}\dot{\theta}+g\cos\theta]
\end{split}
\end{pmatrix}\\
\label{eq:d-chart}
& r_{\TO} = \sqrt{r^2 + x_{\TO}^2-2rx_{\TO}\cos\theta}, \label{eq:rTO}
\end{eqnarray}
\noindent where $x_{\TO}$ is the horizontal distance between the two contact points and $r_{\TO}$ is the length of the back leg.

The event functions are parameterized with the angle of attack and the natural length of the springs.

Switches from the flight phase to the single stance phase are defined by:
\begin{equation}
\mathcal{F}_{ff \rightarrow s}\left(\vec{X}_{ff},\alpha,r_0\right) : \begin{cases}  y - r_0\cos\alpha = 0\\ v_y < 0 \end{cases},
\label{eq:ff2s}
\end{equation}
\noindent which means that the mass is falling and the leg can be placed at its natural length with angle of attack $\alpha$. Therefore, the motion is now defined in the single stance phase. The switch in the other directions is simply:
\begin{equation}
\mathcal{F}_{s \rightarrow ff}\left(\vec{X}_{s},r_0\right) :  r - r_0 = 0.
\end{equation}
\noindent These are the only two event functions involved in the running gait. The map from one phase to the other is defined by:
\begin{equation}
\quad x = -r\cos \theta \qquad  y = r\sin\theta.
\end{equation}
\noindent It is important to have in mind that the origin of the single stance phase is always at the touchdown point.

For the walking gait, we have to consider switches between single and double stance phases:
\begin{equation}
\mathcal{F}_{s \rightarrow d}\left(\vec{X}_{s}, \alpha, r_0\right) : \begin{cases}  r\sin\theta - r_0\cos\alpha = 0\\ \theta > \frac{\pi}{2} \end{cases},
\end{equation}
\noindent which is similar to (\ref{eq:ff2s}) with the additional condition that the mass is tilted forward. Additionally, if we consider the sign of the vertical speed, we differentiate between
walking gait with $v_y$ < 0 and Grounded Running gait with $v_y$ > 0.

The switch from the double stance phase to the single stance phase is defined by:

\begin{equation}
\mathcal{F}_{d \rightarrow s}\left(\vec{X}_{d}, r_0\right) : r_{\TO} - r_0 = 0,
\end{equation}
\noindent with $r_{\TO}$ as defined in (\ref{eq:rTO}). The map from the double stance phase to the single stance phase is the identity. In the other direction we have:
\begin{eqnarray}
&&r_d = r_0 \quad  \theta_d = \alpha, \\
&&x_{\TO} = r_0 \cos\alpha - r_s\cos\theta_s ,
\end{eqnarray}
\noindent where the subscripts indicate the corresponding phase.

If the system falls to the ground ($y \leq 0$), attempts a forbidden transition (e.g. double stance phase to flight phase), or renders $v_x <0$ (motion to the left,``backwards''), we consider that the system fails.

\begin{figure}[htb]
\centering
\includegraphics[width=0.5\textwidth]{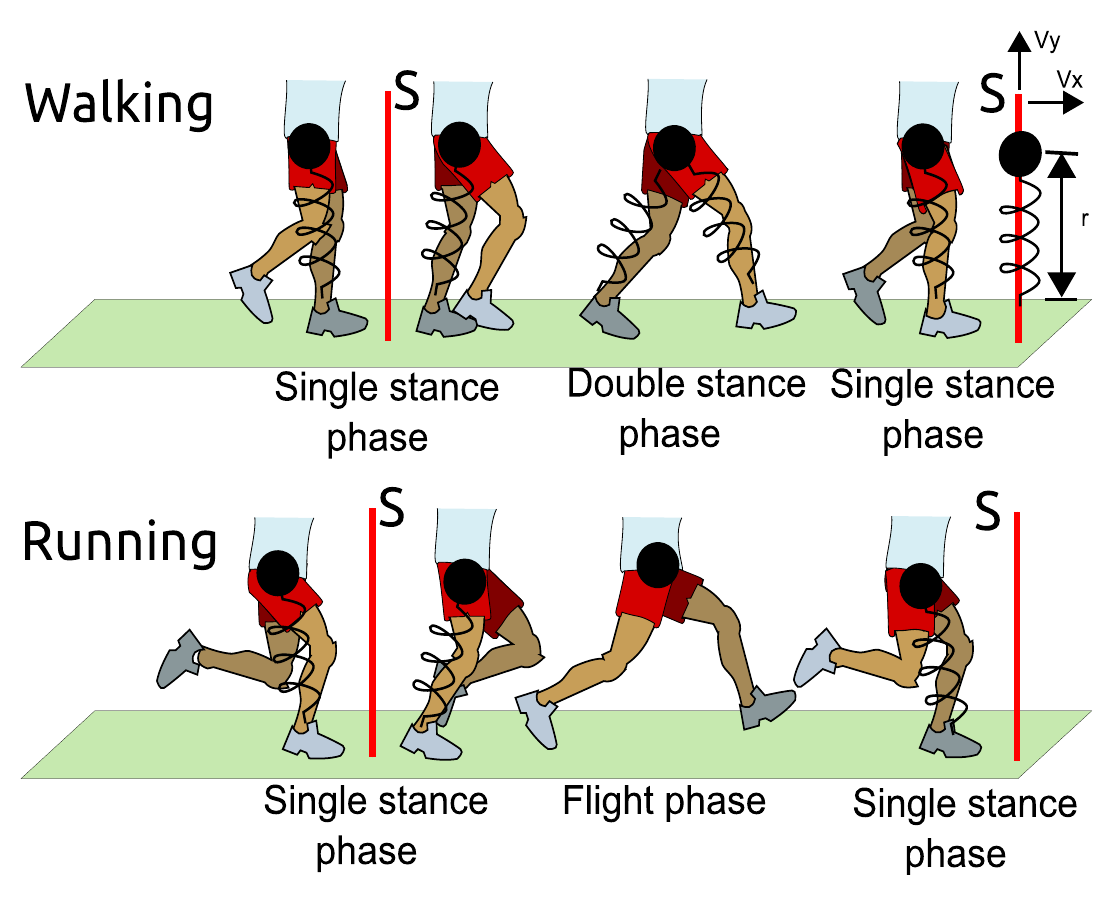}
\caption{\label{fig:SectionS} (Color online) Illustration of the evolution of the SLIP model for running and walking. The different phases are indicated as well as the section S where the system is observed.}
\end{figure}

The state of the model is observed when the trajectory of the system intersects the section $\mathcal{S}$ defined in the single stance phase, i.e. only one leg touching the ground and oriented vertically (Figure~\ref{fig:SectionS}). The results are visualized using the values of the length of the spring $r$ and the radial component of the velocity which, in $\mathcal{S}$, equals the vertical speed $v_y$ (vx is obtained from these values and the equation of constant energy). It is important to note that all possible values of $r$, $v_y$ , and $v_x$ , for a given value of the total energy $E$, lie on an ellipsoid.
\begin{equation}
\label{eq:energy}
E = \frac{1}{2}k\left(r_0-r\right)^2 + \frac{1}{2}m\left(v_x^2 + v_y^2\right) + mgr
\end{equation}

This intermittent observation of the system renders the continuous evolution of the model into a mapping that transforms states in the section at a time $t$, to states in the section at $t+\Delta t$. The interval $\Delta t$ is the time the system takes to reach a new vertical posture, only during periodic gaits it is equivalent to the period of the gait.

\begin{figure}[htb]
\centering
\includegraphics[width=0.9\textwidth]{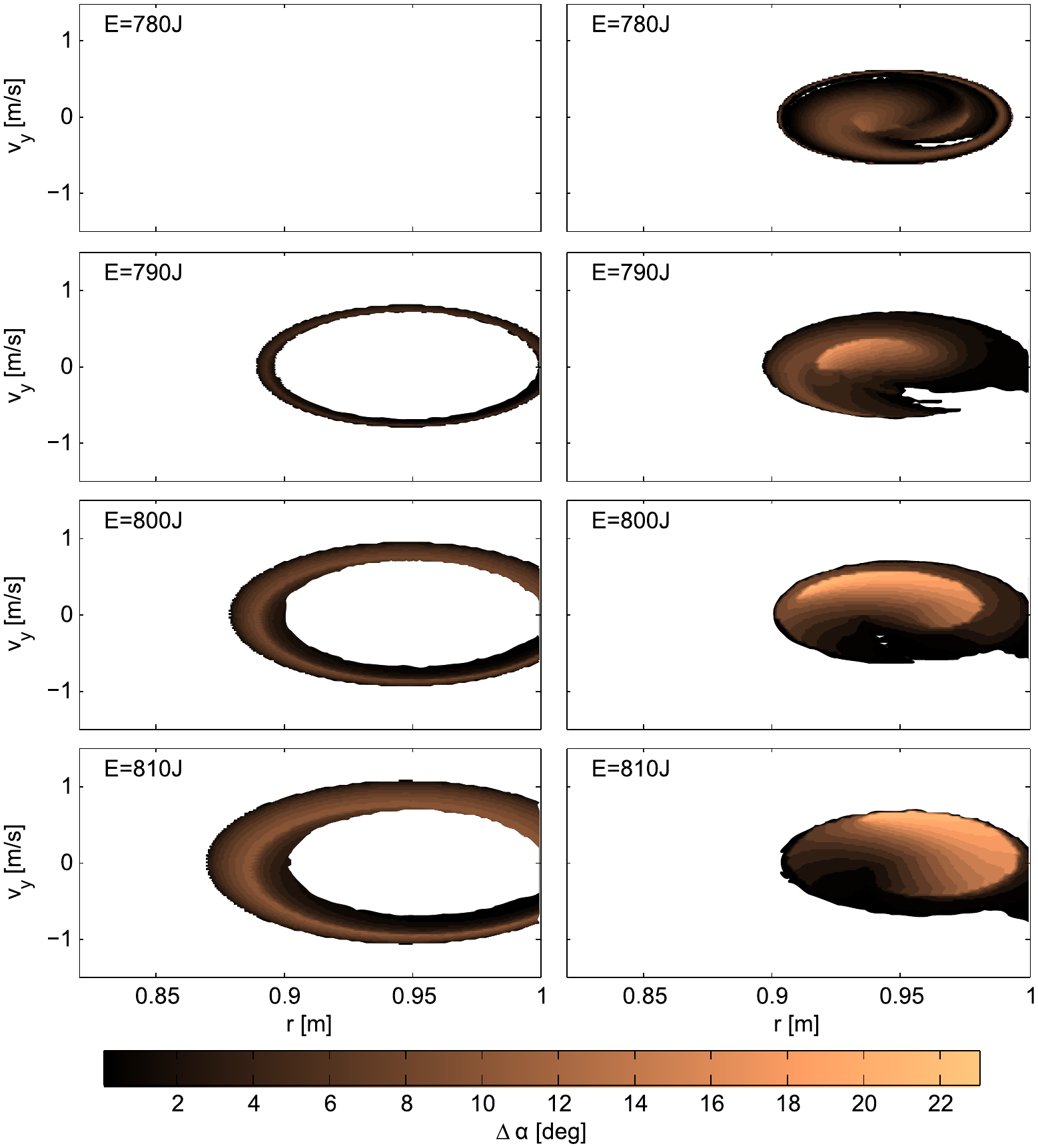}
\caption{\label{fig:Viability3} (Color online) Viability regions for running and walking. The (cooper) gray scale color represents the viability regions for energies between [780\joule-810\joule]. The first column shows the viability region for running and the second column for walking}
\end{figure}

\begin{figure}[htb]
\centering
\includegraphics[width=0.9\textwidth]{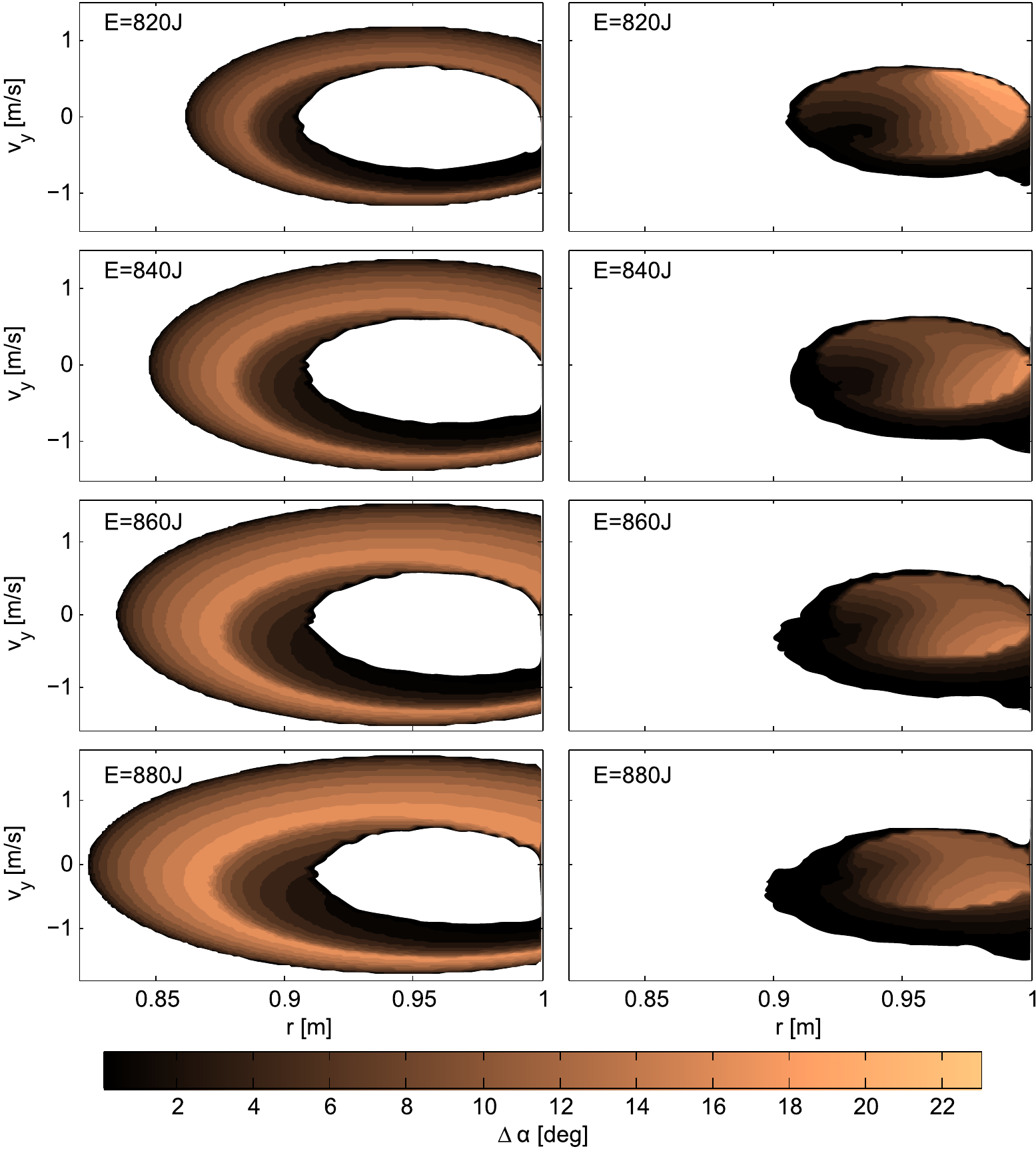}
\caption{\label{fig:Viability4} (Color online) Viability regions for walking and running. The (cooper) gray scale color represents the viability regions for energies between [820\joule-880\joule].The first column shows the viability region for running and the second column for walking}
\end{figure}

Using the maps we calculated the viability regions in the section $\mathcal{S}$. The viability regions are the initial conditions that can perform an step selecting an angle of attack from a continuous interval of length $\Delta \alpha$ the biggest interval size found with the system is $\unit{23} \degree$. Figures~\ref{fig:Viability3}-\ref{fig:Viability4} show different viable regions as a function of the interval length.

\section{Angle of attack estimation from empirical data\label{ap:angleAttkEst}}

In the experimental data of reference~\cite{Ivanenko2011} the angle of the right limb is measure against the vertical. We use this information to estimate the angle of the leg at landing based in two facts. First, the angle of the leg changes more its velocity in the swing phase (the foot is not in contact with ground) than in the support phase (the foot is in contact with the ground), and  second, as soon as the leg changes from the swing phase to the support phase there is a big change of the angular velocity due to the impact of the food against the ground when it lands.

\begin{figure}[htb]
\centering
\includegraphics[width=0.6\textwidth]{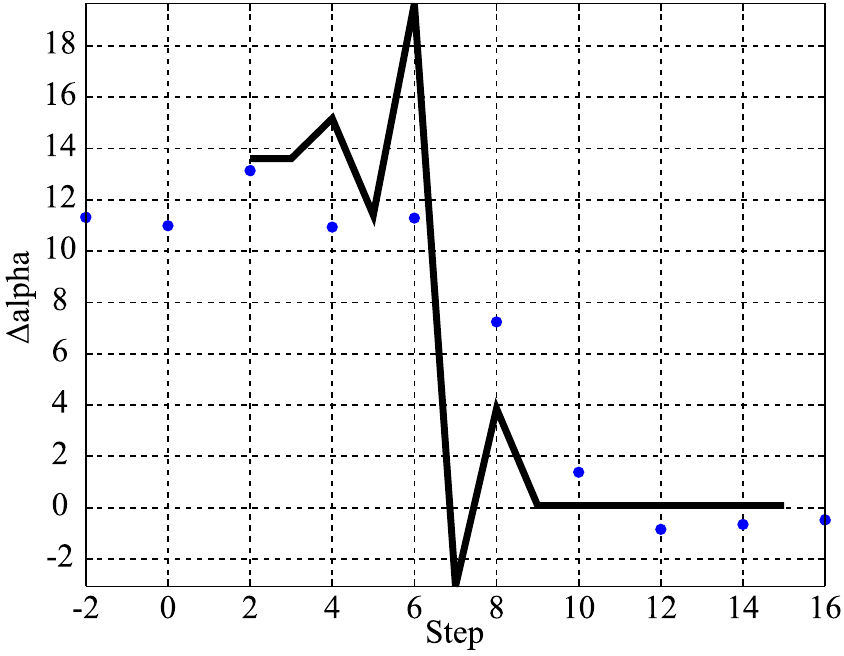}
\caption{\label{fig:angleR2W} (Color online) Change of the angle of attack in the running to walking transition. The solid line represent the change of the angle of attack in the model and the doted line represent the change of the angle of attack in a human experiment. In both case there is a transition from running to walking.}
\end{figure}

The angle of attack identified using this conditions allow the comparison of the strategy in human locomotion and the proposed model. The model qualitatively develops a similar strategy. The difference of the angle of attack between the steady state gait (e.g. walking or running) from the experiment and the model is around five degrees. To facilitate the qualitative comparison of the angle of attack, we evaluate the change of the angle of attack against the angle of attack of walking. Using this measurement, we can avoid the difference of five degrees and focus in the strategy for gait transition.

\begin{figure}[htb]
\centering
\includegraphics[width=0.6\textwidth]{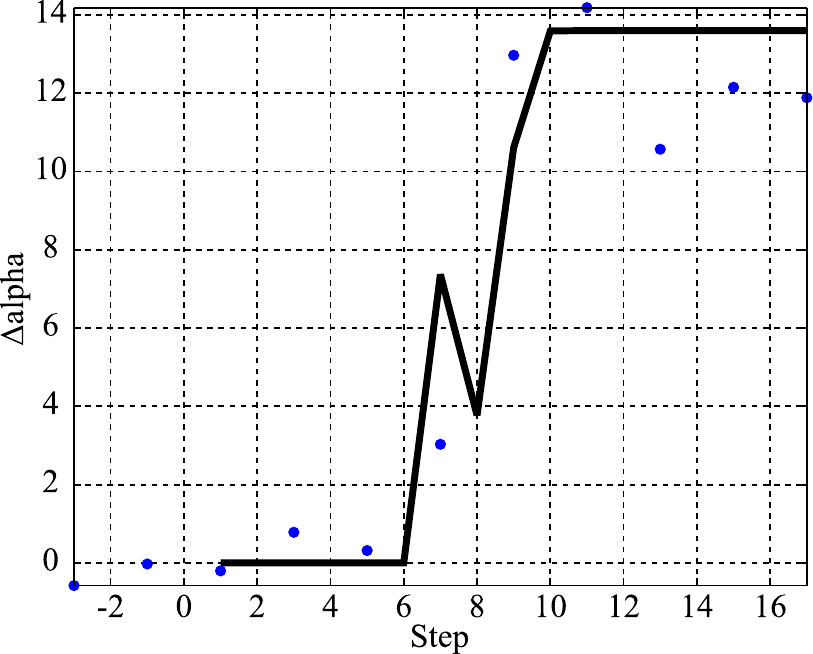}
\caption{\label{fig:angleW2R} (Color online) Change of the angle of attack against in the walking to running transition. The solid line represent the change of the angle of attack in the model and the doted line represent the angle of attack in a human experiment. In both case there is a transition from walking to running.}
\end{figure}

Fig.~\ref{fig:angleR2W} and Fig.~\ref{fig:angleW2R} show that the strategy developed with the model has similar steps and matches the change of the angle of attack in the transition. Fig.~\ref{fig:angleR2W} shows a more drastic change of the angles of attack compare with the experiment result, however the data of the experiment is from one leg which allow the identification of the angle of attack every two steps. This can be emulated with the model selecting only the even or the odd steps. In any of these cases, the change of the angles of attack is going to look less drastic and qualitatively more similar to the ones from the experiment.

\section{Change of phase of hip excursion before and after transition\label{ap:phaseHip}}
\begin{table*}[tbh]
\centering
\begin{tabular}{|l|c|c|}
\hline
Strategy             & $\gait{W} \rightarrow \gait{R}$ & $\gait{R} \rightarrow \gait{W}$ \\
\hline
Const. Froude number &    $\unit{36.3}\degree$         &    $\unit{35.3}\degree$         \\
\hline
Const. hip excursion &    $\unit{55.3}\degree$      &    $\unit{51.5}\degree$      \\
\hline
Fitting experiment   &    $\unit{109.0}\degree$        &    $\unit{110.9}\degree$        \\
\hline
Experiment           &    $\unit{-35.0}\degree$        &    $\unit{86.8}\degree$         \\
\hline
\end{tabular}
\caption{\label{tab:SummPhaseDiff} Change of phases for three strategies and experimental data. None of the transitions shows a phase change in full accordance with the experimental data. The absolute value of the phase change for the transition from walking to running at constant Froude number is very close to the experimental value, however the direction of the change is opposite.}
\end{table*}

\begin{figure}[htb]
\centering
\includegraphics[width=\textwidth]{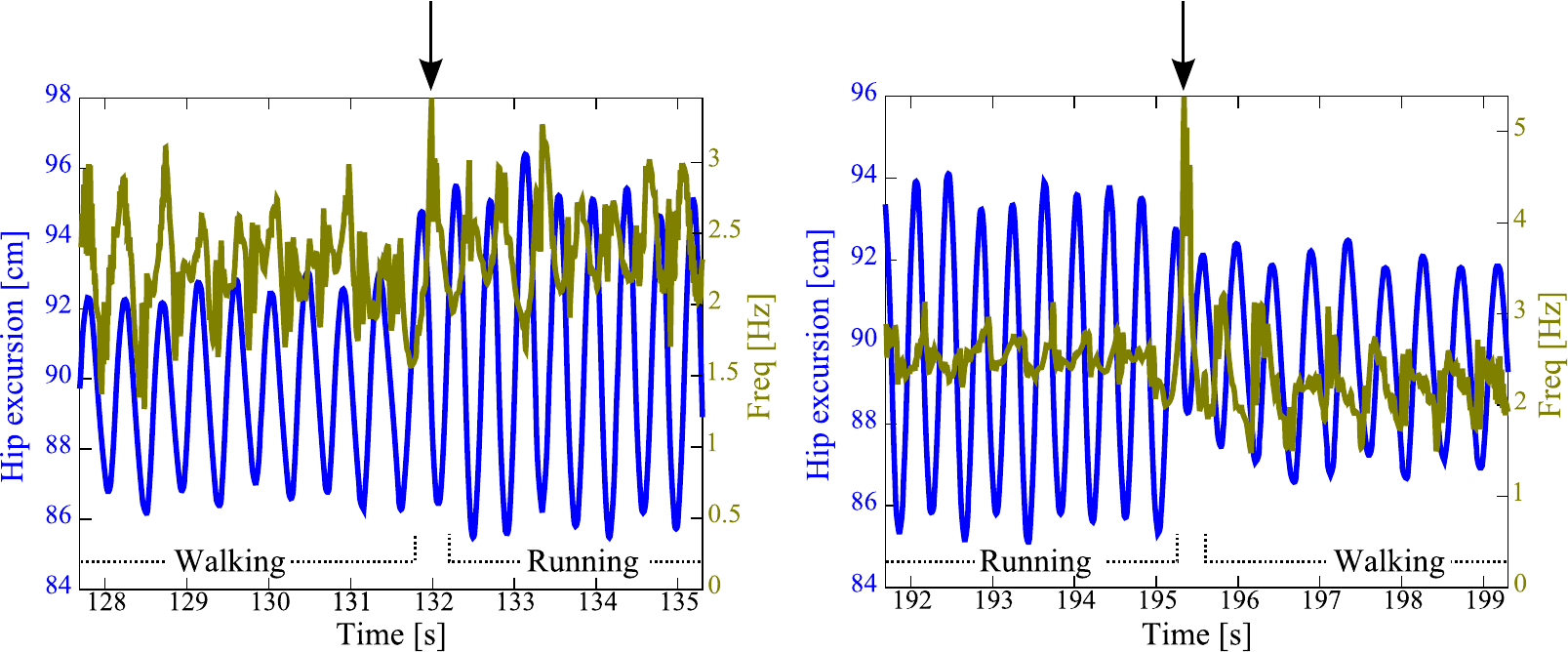}
\caption{\label{fig:TransPoint} (Color online) Transition point determination. Plot of the experimental data (left axis) and the the derivative of the phase signal (right axis). this derivative gives a frequency signal that presents a peak during the transition that is used to determine the transition point (vertical arrow).}
\end{figure}

\begin{figure}[htb]
\centering
\includegraphics[width=0.6\textwidth]{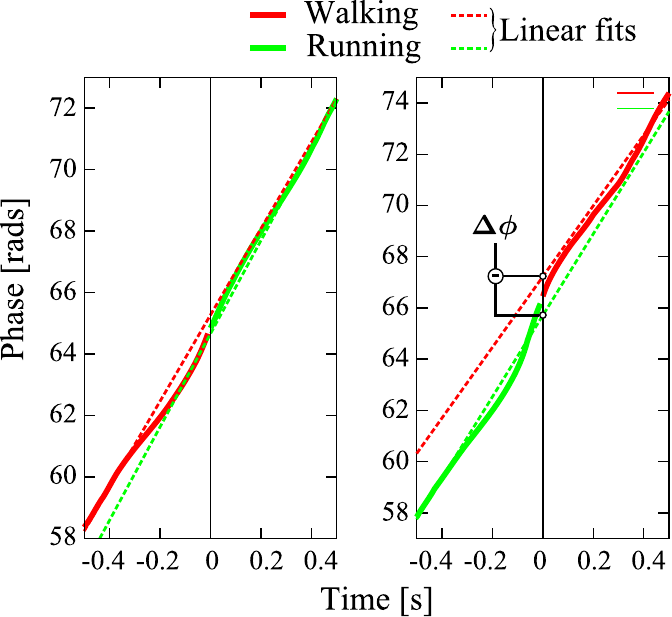}
\caption{\label{fig:PhaseDiff} (Color online) Phase difference calculation. Taking the point of transition as the origin of time, the phase difference is calculate from the intercept of linear fits applied to the two parts of the phase signal. Solid lines show the phase signal for walking and running. Dashed lines show the linear fits.}
\end{figure}

As shown in Figure~\ref{fig:TransPoint} (left axis), during walking and running the hip follows and oscillatory trajectory over time. We compare the phase of these oscillations with respect to the moment of transition. The moment of transition was identified as follows:
\begin{enumerate}
\item Calculate the analytic signal of the hip trajectory by means of the Hilbert transform, e.g. {\tt hilbert} function in GNU Octave's signal package~\cite{octave2002}.
\item Obtain the phase of the signal from the angle of the analytic signal.
\item Take the time derivative of the phase, this is an approximation of the frequency of the oscillations as a function of time.
\item Search for the highest peak in the frequency signal. This point separates the regions of walking from the regions of running.
\end{enumerate}
\noindent Figure~\ref{fig:TransPoint} shows the frequency signal superimposed to the experimental data. The transition point is indicated with a vertical arrow. Taking this point as the origin of time, we calculate the initial phase of walking and the initial phase of running, by means of fitting a first order polynomial to the phase signal of each gait. This is shown in Figure~\ref{fig:PhaseDiff} when applied to the experimental data. The change of phase is calculated as the difference of these initial phases normalized to the interval $(-\pi,\pi]$. The exact same analysis was applied to all the signals, simulated and experimental.

The changes of phase for the three transition strategies presented in the paper are summarized in Table~\ref{tab:SummPhaseDiff}. All the simulated examples are able to match the direction of the change of phase in the running to walking transition. However, none of the transitions shows a phase change in full accordance with the experimental data. The absolute value of the phase change for the transition from walking to running at constant Froude number is very close to the experimental value, however the direction of the change is opposite.

\end{document}